\documentclass[10pt,twocolumn,letterpaper]{article}

\usepackage[pagenumbers]{cvpr} %

\usepackage{booktabs}
\providecommand{\std}[1]{\,{\scriptsize$\pm$#1}}

\AtEndPreamble{%
  \crefname{suppsec}{Supp.~Sec.}{Supp.~Secs.}%
  \Crefname{suppsec}{Supp.~Section}{Supp.~Sections}%
  \crefname{suppfig}{Supp.~Fig.}{Supp.~Figs.}%
  \Crefname{suppfig}{Supp.~Figure}{Supp.~Figures}%
  \crefname{supptab}{Supp.~Tab.}{Supp.~Tabs.}%
  \Crefname{supptab}{Supp.~Table}{Supp.~Tables}%
  \crefname{suppeq}{Supp.~Eq.}{Supp.~Eqs.}%
  \Crefname{suppeq}{Supp.~Equation}{Supp.~Equations}%
}

\definecolor{cvprblue}{rgb}{0.21,0.49,0.74}
\usepackage[pagebackref,breaklinks,colorlinks,allcolors=cvprblue]{hyperref}
\usepackage{tabularx}

\def\confName{CVPR}
\def\confYear{2027}

\title{Vision Pretraining for Efficient Online Reinforcement Learning}

\author{
  Kuo-Han Hung\thanks{Equal contribution. \texttt{\{perryd,khhung\}@stanford.edu}}\quad
  Perry Dong\footnotemark[1]\quad
  Dorsa Sadigh\quad
  Chelsea Finn\\[0.5em]
  \normalfont Stanford University
}

\newcommand{\methodname}{VRP\xspace}

\begin{document}
\maketitle
\begin{abstract}
Visual representations play a central role in the sample efficiency of vision-based online reinforcement learning (RL), yet current methods mostly initialize the encoder from general-purpose vision models or from imitation learning. We find that these pretrained representations do not reliably help online RL: they often perform similarly to or worse than a randomly initialized encoder. In this work, we present Value Representation Pretraining (VRP), a visual representation pretraining method for online RL. VRP combines TD learning with an action-contrastive loss to learn representations that capture task progress and how each action changes the expected future return, information the critic needs during fine-tuning. With the off-policy algorithm EXPO, VRP improves sample efficiency over a randomly initialized encoder by 2.08$\times$ across 3 robomimic and 15 RLBench tasks, whereas encoders pretrained with non-RL pretraining methods like imitation learning, time-contrastive learning, or DINOv2 provide little to no improvement. With the on-policy algorithm PPO, VRP also achieves the highest average success rate on robomimic, raising it from 86.4\% with random initialization to 96.8\%. Website: \href{https://khhung-906.github.io/vrp/}{https://khhung-906.github.io/vrp/}
\end{abstract}
    
\section{Introduction}
Reinforcement learning (RL) fine-tuning has emerged as a promising way to improve pretrained robot policies~\citep{dong2026expoftsampleefficientreinforcementlearning, xu2026rl, dong2026reinforcement}. These policies are typically pretrained with imitation learning (IL) on a small set of task-specific demonstrations, and the same demonstrations remain available during RL fine-tuning. In vision-based control, the choice of visual representation strongly shapes downstream policy learning~\citep{pmlr-v162-parisi22a, pmlr-v270-burns25a, pmlr-v202-hu23h}, and the same holds for RL fine-tuning. Most existing RL fine-tuning methods initialize the visual encoder from imitation pretraining~\citep{ren2024diffusionpolicypolicyoptimization, zhang2026reinflowfinetuningflowmatching, xu2026rl, zang2023behavior}, on the premise that RL can build on the same prior knowledge as the pretrained policy. Yet, how an IL-pretrained encoder compares to other choices for RL fine-tuning remains largely untested. IL only needs to identify the demonstrated action, so its encoder is free to discard information the critic depends on, such as how alternative actions compare or how close a state is to failure. This raises the question: \emph{can we pretrain visual representations that are better suited to RL fine-tuning?}

\begin{figure*}[t]
    \centering
    \includegraphics[width=\textwidth]{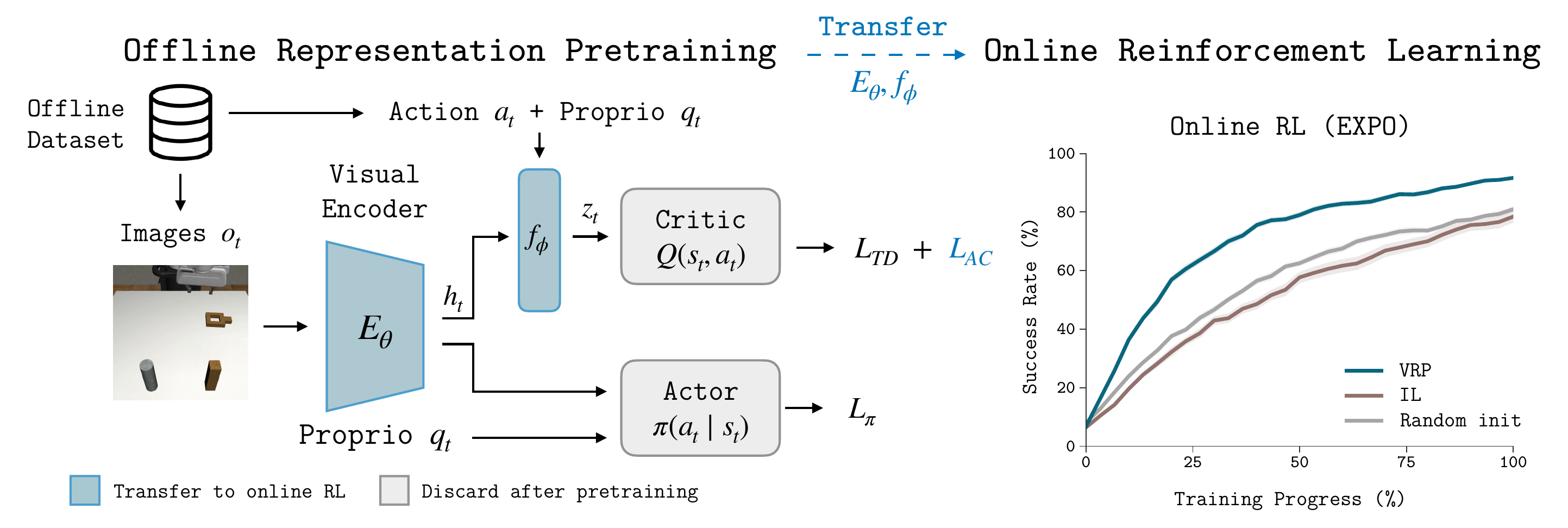}
    \caption{\textbf{Overview of \methodname.} \methodname pretrains visual representations to make online RL more sample efficient. \textbf{Left:} \methodname pretrains the representation with a TD objective and an action-contrastive objective, then transfers the visual encoder and the first critic layer to downstream online RL. \textbf{Right:} Evaluation success rates during online RL with EXPO initialized with \methodname representations versus randomly initialized and imitation-learned representations, averaged across 15 RLBench tasks and 3 robomimic tasks.}
    \label{fig:main}
\end{figure*}

Several approaches offer potential starting points for representation learning in RL. These approaches fall into three categories. The first category consists of encoders learned through imitation learning~\citep{zang2023behavior, xu2026rl}, which are optimized to predict demonstrated actions. As we show in Figure~\ref{fig:main}, these approaches can leave performance on the table when reused for the critic. The second category consists of encoders pretrained on general vision tasks~\citep{pmlr-v139-shah21a, pmlr-v162-parisi22a, xiao2022maskedvisualpretrainingmotor, pmlr-v205-nair23a, NEURIPS2022_548a482d} to capture broadly useful visual features. However, these methods are not trained to reflect how actions affect future outcomes, which is what the critic must predict. Lastly, some works use RL objectives to pretrain the encoder on large multi-task datasets~\citep{kumar2023offline, Kumar-RSS-23} or videos~\citep{10611575, lekkala-vep}, showing that representations learned with offline RL on diverse data can transfer to new tasks. Unlike these works, we consider the common RL fine-tuning setting, where a small set of successful demonstrations for the target task is available. Here, we find that offline RL objectives that train the critic with the TD loss alone learn useful but limited visual representations: because the robot state alone predicts progress on successful demonstrations, the critic fits its target mostly from proprioception and the encoder receives a weak visual signal, a failure we call the \emph{proprioception shortcut}. Transferring the pretrained Q-function itself does not help either: initializing online RL with an offline-trained Q-function gives little benefit over random initialization~\citep{dong2026reallyneedpretrainqfunctions}.

Our key insight is that offline value learning transfers through its representation, not its Q-values: the Q-values grow stale as the policy improves, but features that track task progress and rank actions remain useful. Building on this insight, we present Value Representation Pretraining (\methodname), a method for pretraining visual representations for downstream RL. Starting from the soft actor-critic objective~\citep{haarnoja2018softactorcriticoffpolicymaximum}, which trains a critic with a TD loss and an actor to maximize the critic's value, we augment the objective with an action-contrastive loss that trains the Q-function to separate actions in and out of the demonstration distribution. Specifically, we push up the Q-values of dataset actions and actions sampled from a separately trained imitation learning policy, and push down a soft maximum of the Q-values over all actions, which concentrates on the actions the critic currently overvalues. With these two objectives, the TD loss lets \methodname capture how far the task has progressed and how each action affects future returns, while the action-contrastive loss requires the critic to judge the relative optimality of actions from the visual scene. This forces the encoder to capture task-relevant object information.

Our main contribution is a representation pretraining method for sample-efficient RL fine-tuning. We first show that offline RL on these demonstrations already learns representations that benefit online RL, but that the proprioception shortcut limits this benefit. We then propose \methodname, which avoids this shortcut with an action-contrastive objective and learns representations that capture task-relevant visual information. We evaluate \methodname on 3 robomimic~\citep{robomimic2021} and 15 RLBench~\citep{james2019rlbenchrobotlearningbenchmark} tasks with both an off-policy algorithm, EXPO~\citep{dong2026expo}, and an on-policy algorithm, PPO~\citep{schulman2017proximalpolicyoptimizationalgorithms}. With PPO on robomimic, \methodname raises the average success rate from 86.4\% to 96.8\% relative to a randomly initialized encoder. With EXPO, \methodname improves sample efficiency by $2.08\times$ and raises the average success rate from 81.0\% to 91.7\% over a randomly initialized encoder, whereas non-RL pretraining methods, including imitation learning, time-contrastive learning, and DINOv2, provide little to no improvement.

\section{Related Work}
\textbf{Visual representations for manipulation.} A good visual representation can make learning manipulation policies more data-efficient. One line of work pretrains the vision encoder on large out-of-domain data with standard vision objectives, such as supervised or self-supervised image learning~\citep{pmlr-v162-parisi22a, pmlr-v270-burns25a}, masked autoencoding~\citep{xiao2022maskedvisualpretrainingmotor, pmlr-v205-radosavovic23a}, or predicting the features of masked regions with joint-embedding predictive architectures (JEPA)~\citep{bardes2024revisiting, assran2025vjepa2selfsupervisedvideo}, and then freezes or fine-tunes the encoder for downstream policy learning. Another line of work exploits the temporal structure of video, mostly of humans: time-contrastive learning~\citep{TCN2017, pmlr-v205-nair23a}, video-language alignment~\citep{pmlr-v205-nair23a}, language-conditioned reconstruction and generation~\citep{Karamcheti-RSS-23}, and an implicit goal-conditioned value objective whose embedding distance serves as a reward~\citep{ma2023vip}. A third line pretrains the encoder with dynamics objectives on in-domain demonstrations, training it to predict forward or inverse dynamics from the target domain's own data~\citep{NEURIPS2024_3b8db54b, NEURIPS2023_d36dfcdb}. These objectives are typically evaluated with imitation learning, leaving it unclear whether they transfer to online RL. We instead study which pretraining objectives yield representations that are effective for online visual RL.

\noindent \textbf{Representations for visual RL.} Reinforcement learning from pixels is difficult, since the agent must learn perception and control together from its own interaction, and prior work has approached it in several ways. When demonstrations are available, one approach reuses the backbone of an imitation policy: BPR~\citep{zang2023behavior} pretrains the encoder by imitating the dataset's actions and runs offline RL on top, and RL Token~\citep{xu2026rl} trains a small actor-critic on a compressed embedding of a frozen, imitation-pretrained VLA. Another uses an encoder pretrained on vision tasks, such as a frozen ImageNet-pretrained ResNet~\citep{pmlr-v139-shah21a, NEURIPS2022_548a482d}. A third adds auxiliary losses that give the encoder training signal beyond the reward, such as instance or temporal contrastive learning~\citep{pmlr-v119-laskin20a, pmlr-v139-stooke21a, NEURIPS2023_96d00450, pmlr-v162-mu22a, 9981055}. A related line uses predictive objectives, expecting features that predict the future to also serve control: latent dynamics prediction during RL~\citep{pmlr-v97-gelada19a, schwarzer2021dataefficient}, or latent and inverse dynamics pretrained on offline data~\citep{NEURIPS2021_69eba346}. Finally, the RL objective itself can shape the representation: offline Q-learning or value-based contrastive learning on multi-task data~\citep{kumar2023offline, Kumar-RSS-23, lekkala-vep}, and TD learning on human video~\citep{10611575}, yield representations that transfer to new tasks. Unlike these works, which study how well RL-learned representations transfer across tasks, we show that even a small set of demonstrations for the target task can be used to learn a task-relevant representation that makes online RL more sample-efficient.

\noindent \textbf{Reinforcement learning fine-tuning.} Reinforcement learning  fine-tuning aims to improve a pretrained policy through online interaction. One line of work studies offline-to-online RL~\citep{fql_park2025,hansenestruch2023idqlimplicitqlearningactorcritic,li2026qlearningadjointmatching,dongzheng2026value,pmlr-v306-dong26d,NEURIPS2023_c44a0428,NEURIPS2020_0d2b2061,kostrikov2022offline,NEURIPS2021_a8166da0,NEURIPS2022_d4cc7a2d}, where a critic and a policy are first trained with offline RL and then further improved with online interaction. Another line of work fine-tunes expressive pretrained policies, such as diffusion and flow-matching policies, either with policy gradients through the sampling process or by steering the policy with a learned Q-function~\citep{dong2026expo,li2026qlearningadjointmatching,mcallister2025flowmatchingpolicygradients,zhang2026reinflowfinetuningflowmatching,ren2024diffusionpolicypolicyoptimization}. More recently, some of these methods have been extended to VLAs~\citep{wagenmaker2026posterior,dong2026expoftsampleefficientreinforcementlearning,dong2026reinforcement,xu2026rl}. In this setting, the policy is typically pretrained with imitation learning separately from any value function. A natural way to speed up online learning is to also pretrain the Q-function on offline data, but prior work has shown that this gives little benefit over random initialization~\citep{dong2026reallyneedpretrainqfunctions}. In this work, we find that the useful part of offline value learning is not the Q-values, which become stale as the policy improves, but the representation beneath them. Transferring only this representation, and reinitializing the rest of the agent, substantially improves the sample efficiency of online RL and is complementary to existing fine-tuning algorithms.

\section{Preliminaries}
We formulate the problem as a Markov decision process (MDP) defined by $(\mathcal{S}, \mathcal{A}, \rho, r, \gamma, P, H)$, where $\mathcal{S}$ is the state space, $\mathcal{A}$ is the action space, $\rho(s)$ is the initial state distribution, $P(s' \mid s, a)$ is the transition probability, $r: \mathcal{S} \times \mathcal{A} \rightarrow \mathbb{R}$ is the reward function, $\gamma \in [0, 1)$ is the discount factor, and $H$ is the horizon. Each state $s_t = (o_t, q_t)$ includes an RGB image $o_t$ and the robot proprioceptive state $q_t$. The objective is to learn a policy $\pi(a_t \mid s_t)$ that maximizes the expected discounted return $\mathbb{E}_{\pi}\left[\sum_{t=0}^{H-1} \gamma^t r(s_t, a_t)\right]$.

We consider two stages: offline representation pretraining followed by online RL fine-tuning. We assume access to a fixed offline dataset of transitions from successful demonstrations, $\mathcal{D}_{\text{offline}} = \{(s, a, r, s')\}$, which is used in both stages. During online fine-tuning, we parameterize the Q-function as $Q(s, a) = h_\psi\big(f_\phi(E_\theta(o), q, a)\big)$, where $E_\theta$ is a visual encoder (shared with the policy), $f_\phi$ is the first layer of the critic, including a small proprioception encoder, that fuses the image embedding with proprioception and action, and $h_\psi$ is the output head. During pretraining, a method optimizes an objective $\mathcal{L}_{\text{pre}}$ on $\mathcal{D}_{\text{offline}}$ to learn the representation $E_\theta$, possibly together with auxiliary modules (e.g., $f_\phi$ and a value head, or a decoder). Pretraining methods differ only in $\mathcal{L}_{\text{pre}}$ and in which modules they learn. At the start of online fine-tuning, we transfer $E_\theta$ and, for methods that learn it, $f_\phi$. All remaining parameters, including $h_\psi$ and the policy head, are randomly initialized, and any auxiliary pretraining modules are discarded. The agent then interacts with the environment for $N_s$ steps, storing transitions $(s_t, a_t, r_t, s_{t+1})$ in a replay buffer $\mathcal{D}$ initialized with $\mathcal{D}_{\text{offline}}$, and updates both the policy and the Q-function using $\mathcal{D}$. Our main question is how pretraining $E_\theta$ and $f_\phi$ affects the sample efficiency and final performance of online RL fine-tuning.

\noindent \textbf{EXPO~\citep{dong2026expo}.} We use EXPO as the off-policy RL algorithm in our analysis of representation pretraining. EXPO is a recent RL algorithm that is stable and sample-efficient. EXPO uses two policies: a base diffusion policy $\pi_{\text{base}}$ trained with a supervised loss, and a lightweight edit policy $\pi_{\text{edit}}$. Given a base action $a_t$, the edit policy outputs an edit $\hat{a}_t \in [-\beta, \beta]$, and the edited action is $\tilde{a}_t = a_t + \hat{a}_t$. The edit policy is trained to maximize the Q-value of the edited action with an entropy term weighted by the temperature $\eta$:
{\small
\begin{equation}
\begin{split}
    \mathcal{L}(\pi_{\text{edit}}) = -\mathbb{E}_{(s_t, a_t) \sim \mathcal{D},\, \hat{a}_t \sim \pi_{\text{edit}}} \Big[ & Q(s_t,\, a_t + \hat{a}_t) \\
    & - \eta \log \pi_{\text{edit}}(\hat{a}_t \mid s_t, a_t) \Big]
\end{split}
\label{eq:edit_loss}
\end{equation}
}
With this design, gradients do not need to pass through the base policy, while TD backups still use near-optimal actions. For both inference and TD backups, EXPO uses an on-the-fly (OTF) policy. It samples $N$ base actions, computes their edited versions, and selects the candidate with the highest Q-value:
{\small
\begin{equation}
    \tilde{a}^{*} = \operatorname*{arg\,max}_{a \,\in\, \bigcup_{i=1}^{N} \{a_i,\, \tilde{a}_i\}} Q(s, a),
    \label{eq:otf}
\end{equation}
}
where $Q$ is trained with the TD objective in~\Cref{eq:td_loss}.

\noindent \textbf{PPO~\citep{schulman2017proximalpolicyoptimizationalgorithms}.} In addition to EXPO, we evaluate representation pretraining with PPO, an on-policy policy gradient algorithm. PPO improves the policy with advantages estimated from a learned state-value function $V(s)$, and clips the probability ratio between the new and old policies so that each update stays close to the policy that collected the data. Since PPO trains only on fresh on-policy rollouts and its critic does not take actions as input, we transfer only the visual encoder $E_\theta$.

\section{Value Representation Pretraining}
In this section, we present \methodname, a method for pretraining representations that improve the sample efficiency of online RL. It combines offline value learning with an action-contrastive objective, so that the learned representation captures task-relevant structure and substantially accelerates online learning. We first describe the general procedure we use to pretrain encoders with value-based objectives (\Cref{sec:method-value-pretrain}), then introduce our action-contrastive objective that enables the pretrained representations to distinguish between actions and encode more action-relevant information (\Cref{sec:method-action-contrastive}), and lastly describe implementation details (\Cref{sec:method-impl}).

\subsection{Pretraining Encoders with Value-Based Objectives}
\label{sec:method-value-pretrain}
Our goal is to pretrain the visual encoder offline, from the demonstrations alone, so that before any online interaction it already captures what determines the value of an action. We start from the SAC objective~\citep{haarnoja2018softactorcriticoffpolicymaximum}, a simple way to train an encoder through a value function, and optimize it offline on the demonstration dataset $\mathcal{D}_{\text{offline}}$.

The actor and critic share a visual encoder $E_\theta$ mapping an image $o_t$ to $h_t = E_\theta(o_t)$, each with its own MLP head: the critic takes $(h_t, q_t, a_t)$ and the actor takes $(h_t, q_t)$, where $q_t$ is proprioception and $s_t = (o_t, q_t)$ is the full state. The critic minimizes the TD loss:

{\small
\begin{equation}
    \mathcal{L}_{\mathrm{TD}} = \mathbb{E}_{(s_t, a_t, r_t, s_{t+1}) \sim \mathcal{D}} \Big[ \big( y_t - Q(s_t, a_t) \big)^2 \Big]
\label{eq:td_loss}
\end{equation}
}

where $y_t = r_t + \gamma \bar{Q}(s_{t+1}, \tilde{a}_{t+1}^{*})$ is the TD target, $\bar{Q}$ is a target network, and $\tilde{a}_{t+1}^{*}$ is the next action chosen by the RL policy. The actor minimizes the SAC policy loss:

{\small
\begin{equation}
    \mathcal{L}_\pi = \mathbb{E}_{s\sim \mathcal{D},\ a\sim\pi(\cdot \mid s)}\big[\eta \log \pi(a \mid s) - Q(s,a)\big]
\label{eq:actor_loss}
\end{equation}
}
where $\eta$ is the entropy temperature. Both losses update the shared encoder $E_\theta$.

Our analysis in \Cref{sec:exp-ablation} and \Cref{app:off-rl} shows that value learning alone yields a representation that benefits online RL more than an imitation-pretrained one, though it recovers only part of the gain of \methodname. \Cref{sec:exp-offline-rl} traces this to a proprioception shortcut that arises when the critic is trained with the TD loss alone. We address this in the next section with an additional action-contrastive loss.

\subsection{Action-Contrastive Representation Learning}
\label{sec:method-action-contrastive}
To prevent the critic from fitting its target through the proprioception shortcut, we add an action-contrastive objective that requires the critic to distinguish actions at every state. It must assign higher values to actions drawn from the data distribution than to other actions. Because the correct ranking depends on the scene rather than on proprioception alone, this objective forces the encoder to capture the task-relevant visual information needed for decision making.

We define a good action distribution at each state. The good distribution covers the actions that solve the task: the dataset action $a_{\mathcal{D}}$ and samples from an imitation-trained policy $\pi_{\text{IL}}$ fit to $\mathcal{D}_{\text{offline}}$:

\begin{equation}
    p^{+}(a \mid s) = \tfrac{1}{2}\,\delta_{a_{\mathcal{D}}}(a) + \tfrac{1}{2}\,\pi_{\text{IL}}(a \mid s)
\label{eq:pbc}
\end{equation}

We estimate the soft maximum below by importance sampling from a proposal $p^{-}(a \mid s)$ that mixes uniform random actions and samples of the current policy $\pi$ at both the current state $s$ and the next state $s'$. The action-contrastive loss is:
{\small
\begin{equation}
\begin{split}
\mathcal{L}_{\mathrm{AC}} = \mathbb{E}_{s \sim \mathcal{D}} \Big[\, & \log \sum_{a} \exp Q(s, a) \\
& \;-\; \mathbb{E}_{a \sim p^{+}(\cdot \mid s)} \big[ Q(s, a) \big] \Big]
\end{split}
\label{eq:con_loss}
\end{equation}
}
so the critic is trained to rank the whole good distribution above a soft maximum of $Q$ over all actions, estimated with importance-weighted samples from $p^{-}$ as in CQL~\citep{NEURIPS2020_0d2b2061}. The critic minimizes $\mathcal{L}_Q = \mathcal{L}_{\mathrm{TD}} + \alpha\, \mathcal{L}_{\mathrm{AC}}$. The IL policy $\pi_{\text{IL}}$ is a Gaussian policy trained by maximum likelihood on detached features.

With $p^{+} = \delta_{a_{\mathcal{D}}}$, \Cref{eq:con_loss} reduces to the CQL regularizer. The key difference is the positives. CQL pushes up only the logged action at each state, so with a single positive the critic can satisfy the loss by singling out that one action rather than learning what separates good actions from bad ones. By adding samples from $\pi_{\text{IL}}$ to $p^{+}$, we require the critic to rank a whole region of good actions above a soft maximum over all actions, a ranking that depends on the visual scene rather than on proprioception alone. We also use the objective differently. CQL uses it as a pessimism constraint for offline policy learning, whereas we use it only to pretrain the encoder and the first critic layer, and discard the rest of the agent before online fine-tuning (\Cref{sec:method-impl}). Empirically, pretraining with the standard CQL regularizer, \ie, \methodname without the IL push-up, gives a smaller gain than \methodname (93\% vs.\ 98\% average success rate; \Cref{sec:exp-ablation}), suggesting that a single positive action per state is a weaker signal for representation learning.

\begin{figure*}[t]
    \centering
    \includegraphics[width=\textwidth]{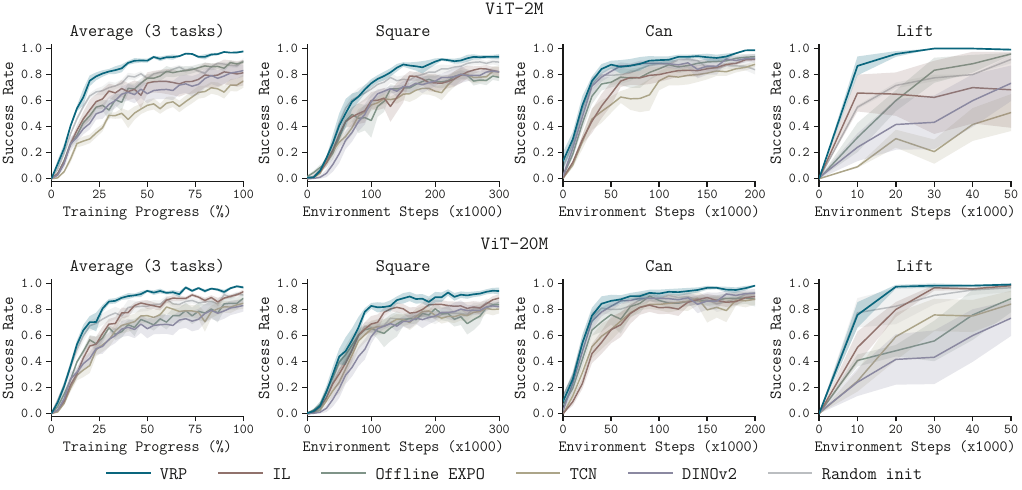}
    \caption{\textbf{Online RL results on robomimic
    (ViT-2M and ViT-20M). }All methods use the same online RL algorithm (EXPO) and differ only
    in the encoder initialization. Curves show the mean over 3 seeds and shaded regions the standard error.}
    \label{fig:exp_results_robomimic}
\end{figure*}

\subsection{Implementation Details} \label{sec:method-impl}

After pretraining, we fine-tune online with EXPO~\citep{dong2026expo}. As shown in \Cref{fig:main}, we do not transfer the whole pretrained agent. We transfer only the visual encoder $E_\theta$ and the first layer of the critic, $f_\phi$, which includes the critic's small proprioception encoder. Its output $z_t = f_\phi(h_t, q_t, a_t)$ is the first point where the critic combines the image, the robot proprioception and the action. The remaining critic layers (which map $z_t$ to $Q(s_t, a_t)$), the actor, and the entropy temperature are initialized from scratch. We test two sizes of the visual encoder $E_\theta$, a ViT~\citep{DBLP:conf/iclr/DosovitskiyB0WZ21}: ViT-2M and ViT-20M, named by their parameter counts (\Cref{tab:vit}). The images from all cameras are encoded together into the embedding $h_t$. We use DrQ-style augmentation~\citep{kostrikov2020image}: 4-pixel edge padding and random cropping back to $84{\times}84$. Every critic and actor head is a 3-layer MLP with 256 hidden units, and only the critic heads use LayerNorm. The online policies follow EXPO's architecture. Both stages use an ensemble of $K = 10$ critic heads as in REDQ~\citep{chen2021randomizedensembleddoubleqlearning}: each head has its own MLP, and the TD target takes the minimum over two heads drawn at random. We transfer $f_\phi$ head by head, so each online critic head starts from its pretrained counterpart. In the offline TD target, $\tilde{a}^{*}_{t+1}$ is the highest-valued of 10 actions sampled from the current policy.

\renewcommand{\topfraction}{0.9}
\renewcommand{\bottomfraction}{0.8}
\renewcommand{\textfraction}{0.07}
\renewcommand{\floatpagefraction}{0.8}
\renewcommand{\dbltopfraction}{0.9}
\renewcommand{\dblfloatpagefraction}{0.8}
\setcounter{topnumber}{3}
\setcounter{bottomnumber}{2}
\setcounter{totalnumber}{5}
\setcounter{dbltopnumber}{3}

\begin{figure*}[t]
    \centering
    \includegraphics[width=\textwidth]{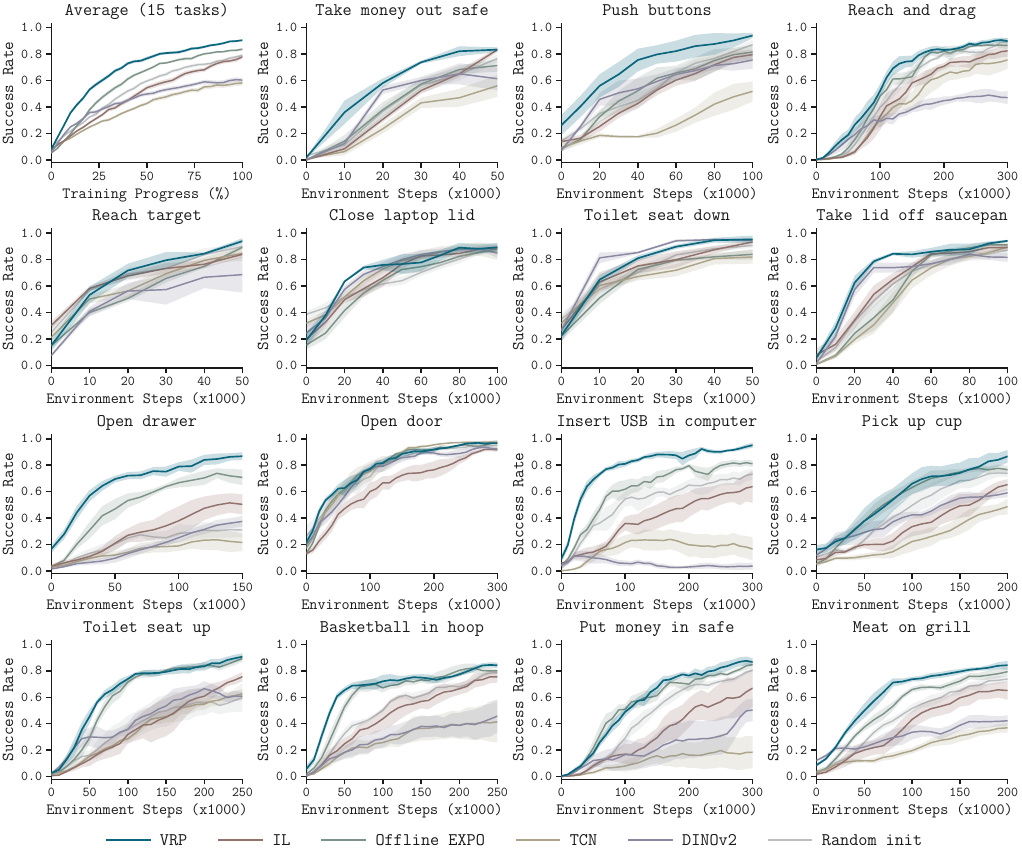}
    \caption{\textbf{Online RL results on RLBench (ViT-2M).} All methods use the same online RL algorithm (EXPO) and differ only in the encoder initialization. Curves show the mean over 3 seeds and shaded regions the standard error.} \label{fig:exp_results_rlbench}
\end{figure*}

\section{Experiments}
\label{sec:experiments}
In this section, we design experiments to evaluate whether representations pretrained with \methodname enable more sample-efficient online RL fine-tuning. We first describe the experimental setup (\Cref{sec:exp-setup}) and then present results organized around the following questions:

\begin{enumerate}[start=1,label={(\bfseries Q\arabic*)}]
    \item Does \methodname improve the sample efficiency and performance of online RL fine-tuning? (\Cref{sec:exp-main})
    \item How does each component of \methodname contribute to its performance? (\Cref{sec:exp-ablation})
    \item Why is offline RL alone not enough? (\Cref{sec:exp-offline-rl})
    \item What do the representations learned by \methodname capture? (\Cref{sec:exp-what-learn})
\end{enumerate}

\subsection{Experimental Setup}
\label{sec:exp-setup}
\textbf{Environments.} We evaluate \methodname on 3 robomimic tasks~\citep{robomimic2021} and 15 RLBench tasks~\citep{james2019rlbenchrobotlearningbenchmark}. All tasks use sparse success rewards and $84\times84$ RGB observations from two cameras (agentview, wrist) in robomimic and four (front, wrist, two shoulders) in RLBench. Learning these tasks efficiently and reliably requires a visual representation that captures task-relevant state while remaining invariant to nuisance variation. For example, the position and orientation of the entire task setup are resampled every episode, so objects appear at different locations, orientations, and scales across episodes. Several tasks also hinge on small visual cues, and some contain visually similar distractors.

\noindent \textbf{Comparisons.} We compare \methodname against several  visual representation pretraining methods:
(i) \textbf{Imitation-pretrained encoder}: a diffusion policy is trained end-to-end with an imitation learning objective on the demonstration data, and its visual encoder is used as the initialization.
(ii) \textbf{Time-contrastive encoder (TCN)}~\citep{TCN2017}: following R3M~\citep{pmlr-v205-nair23a}, we pretrain the encoder on the same demonstrations with R3M's time-contrastive InfoNCE objective, omitting the video-language alignment loss since our data has no language annotations.
(iii) \textbf{DINOv2}~\citep{oquab2024dinov2learningrobustvisual}: we use the smallest released DINOv2 model (ViT-S/14), distilled from ViT-g/14 and pretrained with self-supervised self-distillation on 142M curated images.
(iv) \textbf{Offline EXPO}~\citep{dong2026expo}: to test whether offline value learning alone yields useful representations, we pretrain with the offline EXPO objective and transfer its encoder in the same way as \methodname.
(v) \textbf{Random init}: a randomly initialized encoder with the same architecture as \methodname, trained from scratch together with the policy and value function during online RL fine-tuning.

\subsection{Does \methodname Improve Online RL Fine-Tuning?}
\label{sec:exp-main}
\textbf{Main results.} We first evaluate whether the representation pretrained with \methodname improves the sample efficiency and final performance of online RL fine-tuning. \Cref{fig:exp_results_robomimic} and \Cref{fig:exp_results_rlbench} show the results on 3 robomimic tasks and 15 RLBench tasks. \methodname outperforms all baselines on both benchmarks, improving the average final success rate over a randomly initialized encoder from 90.6\% to 97.8\% on robomimic and from 79.1\% to 90.5\% on RLBench, and reaching the same success rates with $2.08\times$ fewer environment steps on average (\Cref{app:sample_efficiency}). The gains are largest on tasks that require fine-grained perception of task-relevant objects, such as \emph{insert USB in computer} on RLBench and \emph{square} on robomimic.  In contrast, with EXPO, prior pretraining methods do not help: on average, encoders pretrained with IL, TCN, or DINOv2 perform similarly to or worse than an encoder trained from scratch. Offline EXPO, which pretrains the encoder with offline RL alone, supports our claim that offline RL benefits representation learning for online RL but is not enough on its own: it improves over random initialization on RLBench (83.7\% vs. 79.1\%) but falls below it on robomimic (89.7\% vs. 90.6\%), which we trace to the proprioception shortcut in \Cref{sec:exp-offline-rl}. \methodname avoids this shortcut and outperforms offline EXPO on both benchmarks.

\noindent \textbf{Scaling the encoder.} To test whether \methodname remains effective as the backbone grows, we repeat the robomimic experiments with a larger encoder, ViT-20M. As shown in \Cref{fig:exp_results_robomimic}, \methodname continues to outperform the other representation pretraining methods with the larger encoder, reaching higher success rates with fewer environment steps.

\noindent \textbf{Beyond off-policy RL.} In addition to off-policy RL, we also evaluate \methodname with the on-policy policy-gradient method PPO~\cite{schulman2017proximalpolicyoptimizationalgorithms} on the robomimic tasks, keeping everything fixed except the encoder initialization. As shown in \Cref{tab:ppo}, \methodname achieves the highest average success rate (96.8\%, compared with 95.7\% for offline EXPO, 94.0\% for IL, and 86.4\% for random initialization), with the largest gain on \emph{square}, where it reaches 93.8\% while offline EXPO, IL, and random initialization reach 90.4\%, 86.7\%, and 64.2\%, respectively. Offline EXPO is the second-best encoder under PPO, further indicating that offline value learning yields useful representations. These results show that the benefits of \methodname extend from off-policy to on-policy RL fine-tuning.

Additional results and experimental details are in \Cref{app:full} and \Cref{app:exp_details}.

\begin{table}[t]
\centering
\small
\setlength{\tabcolsep}{4pt}
\begin{tabular*}{\columnwidth}{@{\extracolsep{\fill}}lcccc}
\toprule
Encoder init. & Square & Can & Lift & Avg. \\
\midrule
Random init  & 64.2\std{16.5} & 95.4\std{1.5} & 99.6\std{0.4} & 86.4 \\
IL           & 86.7\std{5.2} & 95.4\std{1.1} & 100.0\std{0.0} & 94.0 \\
Offline EXPO & 90.4\std{1.5} & 97.9\std{0.4} & 98.8\std{0.7} & 95.7 \\
TCN          & 75.4\std{5.3} & 95.4\std{2.2} & 97.9\std{2.1} & 89.6 \\
DINOv2       & 81.3\std{2.8} & 92.1\std{1.8} & 98.3\std{1.1} & 90.6 \\
\midrule
\methodname  & 93.8\std{2.9} & 97.1\std{1.1} & 99.6\std{0.4} & \textbf{96.8} \\
\bottomrule
\end{tabular*}
\caption{\textbf{Online RL fine-tuning with PPO.} Final success rate (\%) on robomimic (mean $\pm$ s.e. over 3 seeds). All methods use the same PPO setup and differ only in the encoder initialization.}
\label{tab:ppo}
\end{table}

\subsection{Ablation Studies}
\label{sec:exp-ablation}
\begin{figure}[tb]
    \centering
    \includegraphics[width=\columnwidth]{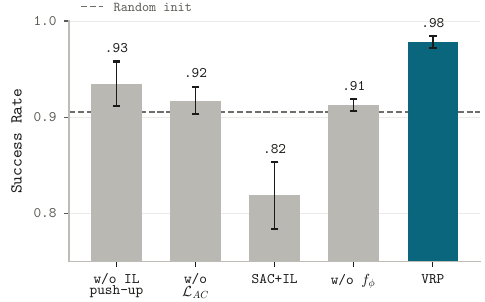}
    \caption{\textbf{Ablation studies.} Final success rate averaged over \emph{square}, \emph{can}, and \emph{lift}. The dashed line is training from scratch. Removing or replacing any component lowers performance.}
    \label{fig:ablation}
    \vspace{-10pt}
\end{figure}

We conduct ablation studies to evaluate the contribution of each component in \methodname. We compare \methodname with four variants: (1) removing the IL push-up in \Cref{eq:pbc}; (2) removing $\mathcal{L}_{\mathrm{AC}}$ in \Cref{eq:con_loss}; (3) using only the TD loss for critic training, with IL as an actor regularizer; and (4) omitting the pretrained weights of the critic's first MLP layer $f_\phi$. The results are shown in \Cref{fig:ablation}. First, removing the IL push-up in \Cref{eq:pbc} lowers the average success rate from 0.98 to 0.93. Without it, $\mathcal{L}_{\mathrm{AC}}$ raises Q-values only at dataset actions, which likely provides the encoder with a weaker signal for the object--gripper relation captured by \methodname (\Cref{sec:exp-what-learn}). Removing $\mathcal{L}_{\mathrm{AC}}$ results in a similar drop, to 0.92, close to training from scratch (0.91). Without $\mathcal{L}_{\mathrm{AC}}$, the critic is trained using the TD loss alone. As we show in \Cref{fig:shortcut}, such a critic predicts values primarily from proprioception and largely ignores visual inputs, so its encoder learns little beyond a randomly initialized one. We also evaluate SAC+IL, which adds an IL term to the actor objective instead of regularizing the critic. It performs even worse than random initialization (0.82 vs.\ 0.91); as \Cref{sec:exp-offline-rl} shows, its critic relies almost entirely on proprioception, leaving the encoder with little visual signal. Finally, randomly initializing the first critic layer $f_\phi$ while still transferring the pretrained encoder lowers the success rate to 0.91, on par with training from scratch. This indicates that \methodname also learns useful information in $f_\phi$ about how states and actions relate, and that transferring the encoder together with $f_\phi$ is important for downstream RL.

\subsection{The Proprioception Shortcut in Offline RL}
\label{sec:exp-offline-rl}

\begin{figure}[tb]
    \centering
    \includegraphics[width=\columnwidth]{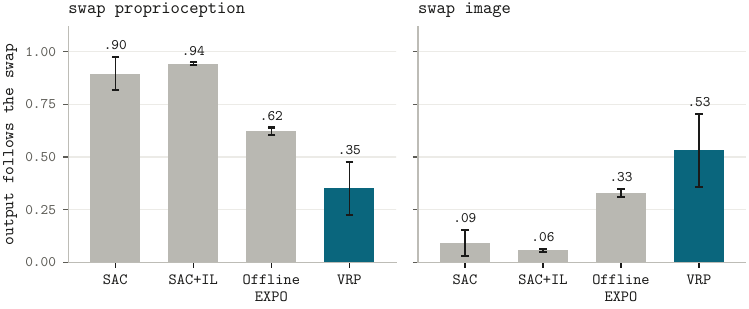}
    \caption{\textbf{Proprioception shortcut.} Offline RL critics trained on successful demonstrations rely mostly on proprioception and largely ignore the image, while VRP uses both.}
    \label{fig:shortcut}
    \vspace{-10pt}
\end{figure}

\begin{figure*}[t]
    \centering
    \includegraphics[width=\textwidth]{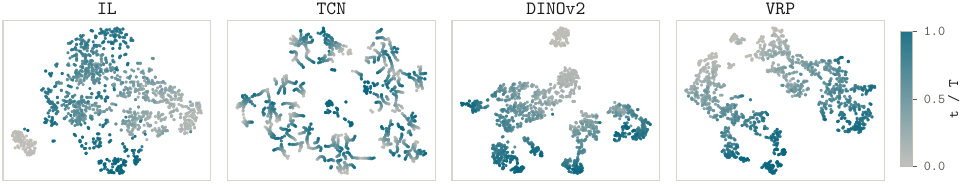}
    \caption{\textbf{t-SNE of learned representations.} t-SNE embeddings of features from different representations on robomimic \emph{square}. Each dot is a single observation, colored by its timestep within the episode; darker dots correspond to later timesteps.}
    \label{fig:tsne}
\end{figure*}

\begin{figure*}[t]
    \centering
    \includegraphics[width=\textwidth]{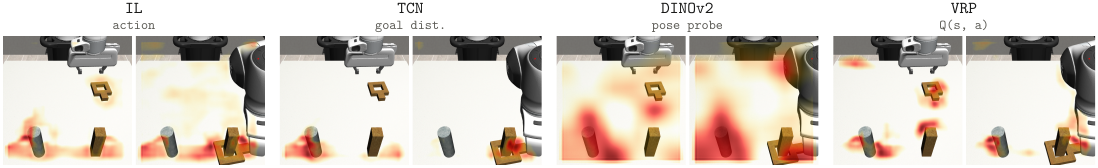}
    \caption{\textbf{Grad-CAM.} Grad-CAM saliency maps on the agentview camera in \emph{square} for different encoders, computed with respect to each encoder's target (IL: predicted action; TCN: distance to the goal embedding; DINOv2: an object pose probe; \methodname: $Q(s,a)$).}
    \label{fig:grad_cam}
\end{figure*}

As shown in~\Cref{sec:exp-ablation} and~\Cref{app:off-rl}, the offline RL objective alone provides only part of the benefit of \methodname. We attribute this to a phenomenon we call the \emph{proprioception shortcut}: when the critic is trained with the TD loss alone on successful demonstrations, it predicts value heavily from the proprioceptive state $q$ and largely ignores the visual input, giving the image encoder little learning signal. To quantify this, we measure how much Q-functions trained with different objectives (offline EXPO, SAC, SAC+IL, and \methodname) rely on $q$ versus the image latent $E_\theta(o)$ in \Cref{fig:shortcut}. We pair each demonstration state $i$ with a random state $j$, replace only $q$ (or only $E_\theta(o)$) with $j$'s, regress the resulting change in Q onto $Q_j - Q_i$ over all pairs, and report the least-squares slope (1: Q fully follows that input; 0: Q ignores it; \Cref{app:proprio}). The SAC and SAC+IL critics follow $q$ almost entirely ($0.90$, $0.94$) and barely respond to the image ($0.09$, $0.06$), and offline EXPO still relies mainly on $q$ ($0.62$ vs.\ $0.33$). Because the robot state alone already predicts task progress on demonstrations, these critics can fit the value without learning visual features. In contrast, \methodname avoids this shortcut through the action-contrastive loss $\mathcal{L}_{\mathrm{AC}}$: its Q-function relies more on the image than on proprioception ($0.53$ vs.\ $0.35$), which yields better visual representations.

\begin{figure}[tb]
    \centering
    \includegraphics[width=\columnwidth]{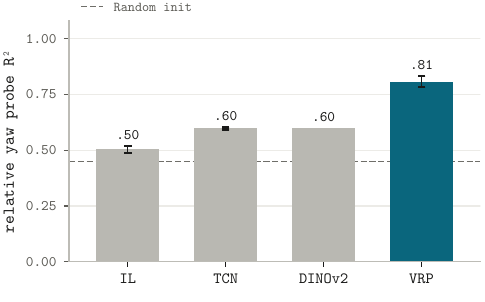}
    \caption{\textbf{Object--gripper orientation probing.} $R^2$ of linear probes predicting the object's yaw relative to the gripper on robomimic \emph{square}, evaluated on held-out demonstrations. Dashed line: randomly initialized encoder.}
    \label{fig:probe}
    \vspace{-10pt}
\end{figure}

\subsection{What Does \methodname Learn?}
\label{sec:exp-what-learn}
To understand why \methodname helps online RL, we analyze the learned representations on robomimic \emph{square}: how they organize observations over a task, and whether they capture task-relevant object state.

\noindent \textbf{t-SNE~\citep{JMLR:v9:vandermaaten08a} analysis.}
\Cref{fig:tsne} shows t-SNE embeddings of demonstration observations, colored by normalized timestep $t/T$. In the IL embedding, observations from different stages of the task are interleaved, suggesting that predicting actions does not require organizing states by task progress. DINOv2 forms clusters that do not follow task stage, consistent with its generic image-level objective. TCN produces elongated structures that largely separate individual trajectories, so the same task stage from different demonstrations is not grouped together. In contrast, \methodname arranges observations along a smooth progression from early to late timesteps shared across episodes. We attribute this to value learning: since the value of a state increases with task progress, states at the same stage of different episodes tend to be mapped to nearby features.

\noindent \textbf{Grad-CAM~\citep{Selvaraju_2017_ICCV} and object probes.}
We next examine how well each encoder captures information about the task-relevant object. \Cref{fig:grad_cam} shows where each encoder attends, and \Cref{fig:probe} reports how well a linear probe on its features predicts the object's yaw relative to the gripper. Attending to the object alone is not sufficient. The IL encoder attends to the object, yet its probe reaches only $R^2 = 0.50$, barely above a randomly initialized encoder ($0.45$). DINOv2 spreads attention across every object in the scene and reaches $0.60$, as its features are not specialized to the task-relevant object. TCN also reaches $0.60$, likely because its features mainly separate individual demonstrations rather than encode how the object relates to the gripper. In contrast, \methodname concentrates on the object and its contact with the gripper, and achieves the highest $R^2$ ($0.81$). Such object-centric information is what the critic needs to judge which actions lead to success, giving online RL a better starting point than the other encoders.
\section{Discussion}
We presented \methodname, a visual representation pretraining method for online RL fine-tuning. By pairing TD learning with an action-contrastive objective, \methodname avoids the proprioception shortcut that arises when the critic is trained with the TD loss alone, and learns representations aligned with task progress and object information. These representations give the online critic a strong starting point, which leads to more sample-efficient fine-tuning across both off-policy and on-policy RL algorithms.

While these results are promising, \methodname{} has several limitations. It pretrains a separate encoder for each task from that task's demonstrations, so it does not yet produce a general visual representation that transfers across tasks; extending the objective to multi-task data is a natural next step. The action-contrastive loss also assumes that the offline data consists of successful demonstrations, and its behavior on suboptimal or mixed-quality data remains to be studied. Finally, our experiments are limited to simulation; evaluating \methodname{} on real robots and with large pretrained backbones such as VLAs is an important direction for future work.

\section*{Acknowledgment}

This work was supported in part by NSF CAREER, NSF \#1941722, RAI Institute, ONR grant N00014-22-1-2293, and ONR grant N00014-22-1-2621. We thank Google Cloud for
providing TPU resources and credits.

{
    \small
    \bibliographystyle{ieeenat_fullname}
    \bibliography{main}
}

\clearpage
\clearpage
\maketitlesupplementary

\appendix
\crefalias{section}{suppsec}
\crefalias{subsection}{suppsec}
\crefalias{subsubsection}{suppsec}
\crefalias{figure}{suppfig}
\crefalias{table}{supptab}
\crefalias{equation}{suppeq}
\section{Full Experiment Results} \label{app:full}

\begin{figure*}[!t]
    \centering
    \includegraphics[width=\textwidth]{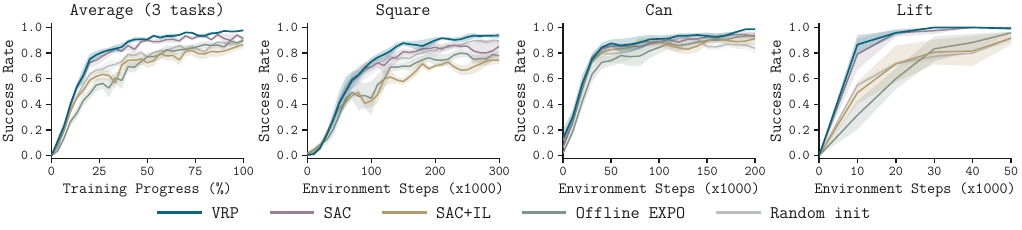}
    \caption{\textbf{Online RL on robomimic with encoders pretrained by different offline RL objectives.} All methods use the same online RL algorithm (EXPO) and differ only in encoder initialization. On these tasks, most offline RL methods yield limited improvement compared to random initialization. Curves show the mean over 3 seeds and shaded regions the standard error.}
    \label{fig:or2o}

    \vspace{0.8em}
    \includegraphics[width=\textwidth]{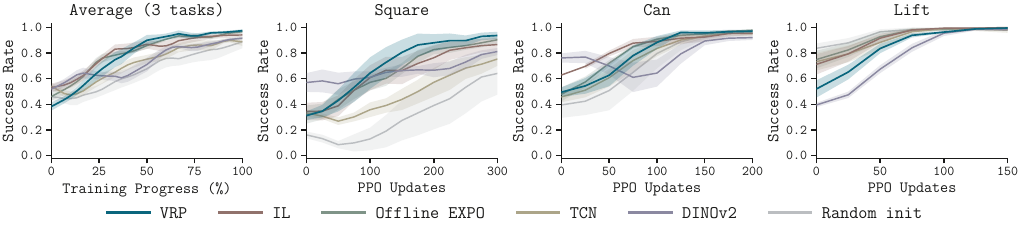}
    \caption{\textbf{Online RL results on robomimic with PPO.} All methods use the same PPO setup and differ only in the encoder initialization. Curves show the mean over 3 seeds and shaded regions the standard error.}
    \label{fig:ppo_full}
\end{figure*}

\subsection{Offline RL Representations for Online RL} \label{app:off-rl}

\Cref{sec:exp-ablation} shows that pretraining the encoder with value learning alone recovers only part of the gain of \methodname. To check whether this holds for standard offline RL objectives in general, we pretrain the encoder with SAC, SAC+IL (SAC with an imitation term added to the actor loss), and offline EXPO on the same demonstrations. We then  run the same online RL algorithm for every method. As shown in \Cref{fig:or2o}, none of these objectives matches \methodname. SAC speeds up early learning compared to a randomly initialized encoder but converges to a similar final success rate, while offline EXPO does not improve over random initialization on these robomimic tasks, even though it does on RLBench (\Cref{sec:exp-main}) and under PPO (\Cref{tab:ppo}). SAC+IL learns more slowly than random initialization and ends at a lower success rate, consistent with its result in the ablation (\Cref{sec:exp-ablation}). The gap is largest on \emph{square}, which requires precise spatial information about the nut and the peg. This matches the proprioception shortcut analyzed in \Cref{sec:exp-offline-rl}. When trained only on successful demonstrations, these critics can fit the value from the robot state alone, so the encoder gets little signal to learn the visual features that precise manipulation needs. \methodname avoids this shortcut with the action-contrastive loss, which makes the critic, and therefore the encoder, depend on the image.

\subsection{Full PPO Plots}
We show the full learning curves of PPO fine-tuning on the three robomimic tasks in \Cref{fig:ppo_full}, which complement \Cref{tab:ppo} in the main paper. Unlike with EXPO, every pretrained encoder improves over random initialization under PPO. One possible reason is that in PPO the encoder is also trained by the policy loss (\Cref{sec:hp}), so features that predict demonstrated actions are directly useful to the actor. In EXPO, the encoder is trained only by the critic loss, which needs information about how actions affect future return. \methodname is the only initialization that helps in both settings.

\section{Additional Analysis}
\subsection{Full Grad-CAM Results}
\begin{figure*}[t]
    \centering
    \includegraphics[width=\textwidth]{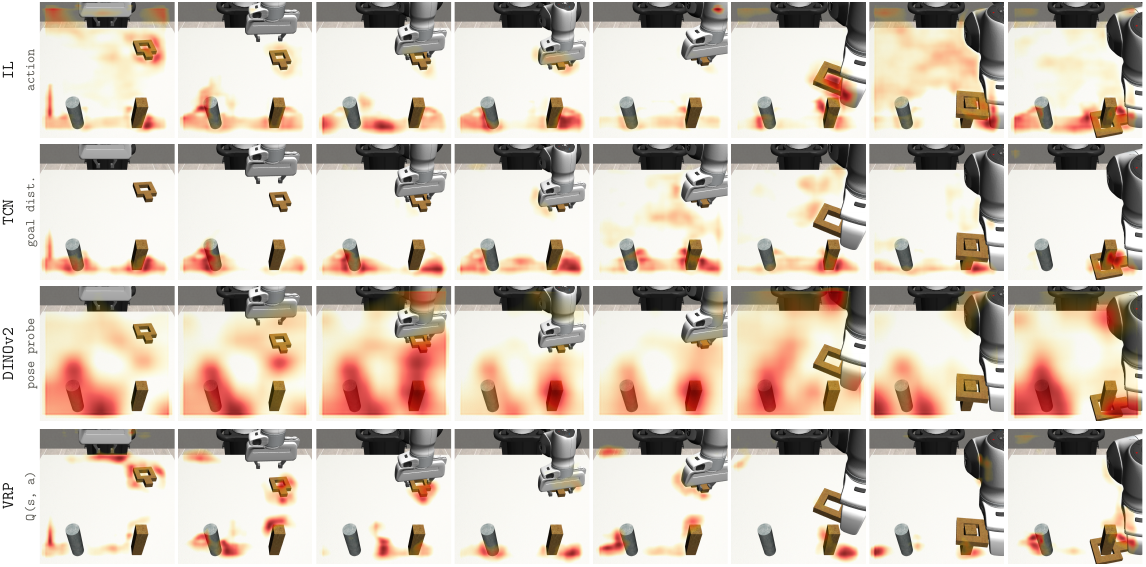}
    \caption{\textbf{Grad-CAM.} Full Grad-CAM saliency maps on the agentview camera in robomimic \emph{square} for different encoders, computed with respect to each encoder's target (IL: predicted action; TCN: distance to the goal embedding; DINOv2: an object pose probe; \methodname: $Q(s,a)$).}
    \label{fig:full_grad_cam}
\end{figure*}

We present the full Grad-CAM visualizations for different visual encoders over a complete episode in \Cref{fig:full_grad_cam}. Throughout the episode, the representations learned by \methodname focus on objects relevant to task completion, shifting from the square nut at the start to the peg during the aiming stage. In contrast, other methods either fail to focus on these objects or distribute their attention across task-irrelevant regions.

\subsection{Proprioception Swap Test} \label{app:proprio}
We train each critic (offline EXPO, SAC, SAC+IL and \methodname) offline for 300k steps on the 200 \emph{square} demonstrations with the ViT-2M encoder, using 3 seeds. Let $z = E_\theta(o)$. The critic output is $y = Q(z, q, a)$ at the dataset action, averaged over the 10 Q-heads. We use every state from the first 50 training demonstrations, each truncated at its first reward. We randomly permute the states so that each state $i$ is paired with a state $j$, and compute the original output $y_i = Q(z_i, q_i, a_i)$, the output of the full partner state $y_j = Q(z_j, q_j, a_j)$, and the output with one input swapped: $y^{q}_i = Q(z_i, q_j, a_i)$ for proprioception and $y^{z}_i = Q(z_j, q_i, a_i)$ for the image. For $c \in \{q, z\}$, we fit a single least-squares slope over all pairs,
\begin{equation*}
\mathrm{follow}_c = \frac{\sum_i (y^c_i - y_i)(y_j - y_i)}{\sum_i (y_j - y_i)^2},
\end{equation*}
rather than averaging per-pair ratios, so that pairs with a small $y_j - y_i$ cannot dominate. A value of 1 means that swapping only that input moves Q all the way to the partner's value, and 0 means that Q ignores it. We compute this once per seed and report the mean $\pm$ s.e. over seeds (\Cref{fig:shortcut}). The two values need not sum to 1: $y_j$ also uses the partner's action, and the joint effect of $z$ and $q$ on Q is not attributed to either input alone.

\section{Measuring Sample Efficiency}
\label{app:sample_efficiency}

Let $S^k_m(t)$ be the success rate of method $m$ on task $k$ after $t$ environment steps (mean over 3 seeds, linearly interpolated between evaluations), and let $\tau^k_m(y)$ be the first step at which $S^k_m(t) \ge y$. For each task, we take 400 evenly spaced success levels $\mathcal{Y}_k$ between where both curves start and the final success rate of Random init, and average the ratio of steps needed to reach each level:
\begin{equation}
  \rho_k = \frac{1}{|\mathcal{Y}_k|} \sum_{y \in \mathcal{Y}_k}
           \frac{\tau^k_{\text{Random init}}(y)}{\tau^k_{\mathrm{VRP}}(y)} .
\end{equation}
Levels that \methodname{} never reaches are excluded. We aggregate over the 18 tasks with the geometric mean, $\bar{\rho} = \exp\big(\frac{1}{18}\sum_k \ln \rho_k\big)$, which treats speed-ups and slow-downs symmetrically and limits the influence of a single task with a large ratio.

\section{Experiment Details} \label{app:exp_details}
\subsection{Benchmarks and Evaluation}

We evaluate on three robomimic tasks~\citep{robomimic2021} and fifteen RLBench tasks~\citep{james2019rlbenchrobotlearningbenchmark}. All of them are vision-based manipulation tasks with a single Franka Panda arm and a sparse success reward. The policy observes RGB images from several cameras together with low-dimensional proprioception. It never observes privileged object states. All methods use the same demonstrations, the same environment interface and the same evaluation protocol on a given task.

\begin{figure}[t]
    \centering
    \includegraphics[width=\columnwidth]{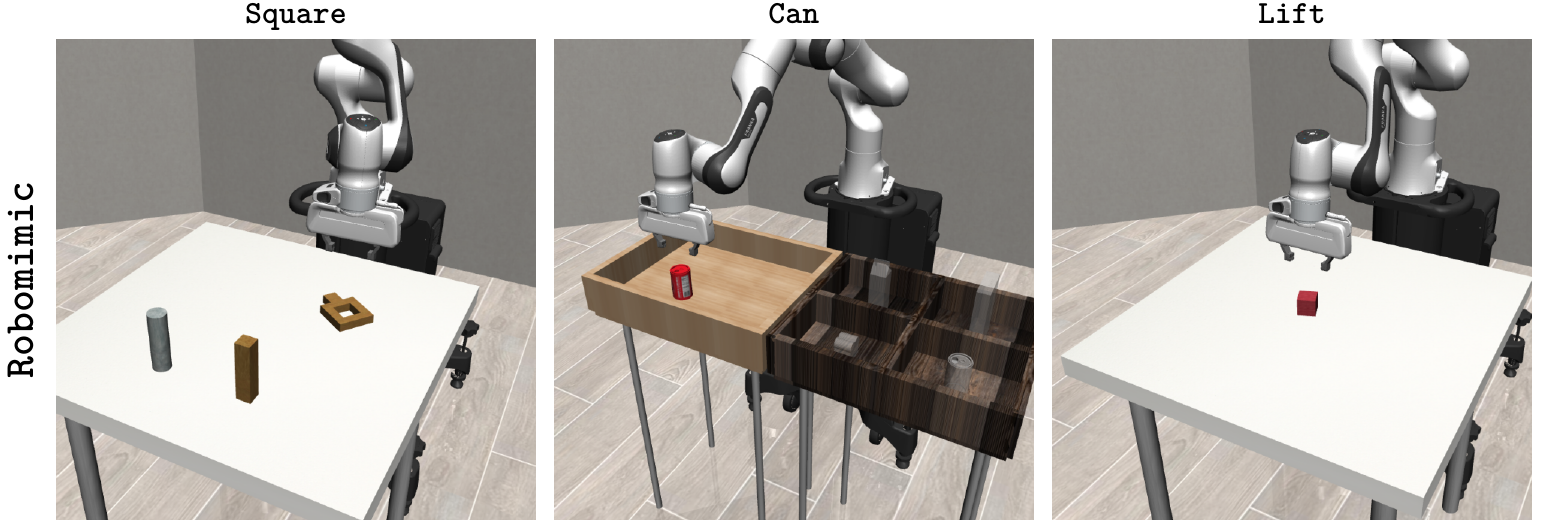}
    \caption{Full visualization of the robomimic tasks we tested.}
    \label{fig:robomimic}
\end{figure}

\paragraph{robomimic~\citep{robomimic2021}.}
We use three tasks from robomimic. They are simulated in MuJoCo through robosuite. \Cref{tab:robomimic_tasks} lists the tasks, and \Cref{fig:robomimic} shows them.

\emph{Observations.} Each observation has two $84\times84$ RGB images: one from the third-person \texttt{agentview} camera, which faces the robot across the table, and one from the wrist camera (\texttt{robot0\_eye\_in\_hand}). It also has 9-D proprioception: the end-effector position (3) and quaternion (4), and the two gripper finger positions (2).
\emph{Actions.} The action is 7-D: a delta end-effector pose (3-D translation and 3-D axis-angle rotation) tracked by robosuite's operational-space controller (OSC\_POSE) at 20\,Hz, plus a gripper open/close command.

\emph{Demonstrations.} We use robomimic's human teleoperation datasets in their image form. We subsample a fixed random subset of demonstrations for each task (\Cref{tab:robomimic_tasks}). All methods and all seeds share the same subset. PH denotes the proficient-human dataset and MH the multi-human dataset.

\begin{table}[h]
\centering
\small
\caption{The three robomimic tasks, the demonstrations used for each, and their online interaction budgets. PH: proficient-human; MH: multi-human.}
\label{tab:robomimic_tasks}
\begin{tabular}{@{}p{0.12\linewidth}p{0.47\linewidth}rr@{}}
\toprule
Task & Goal and scene & Demos & \shortstack[r]{Online\\steps} \\
\midrule
Lift   & Pick up a small cube from the table and lift it above a height threshold. The cube's position and orientation on the table are randomized. & 5 PH & 50k \\
Can    & Pick a soda can out of a bin and place it in the matching compartment of a second bin. The can's initial pose in the source bin is randomized. & 100 MH & 200k \\
Square & Pick up a square nut by its handle and fit it onto a square peg. The nut's position and yaw are randomized. The task needs precise alignment of the nut with the peg. & 200 PH & 300k \\
\bottomrule
\end{tabular}
\end{table}

\begin{figure*}[t]
    \centering
    \includegraphics[width=\textwidth]{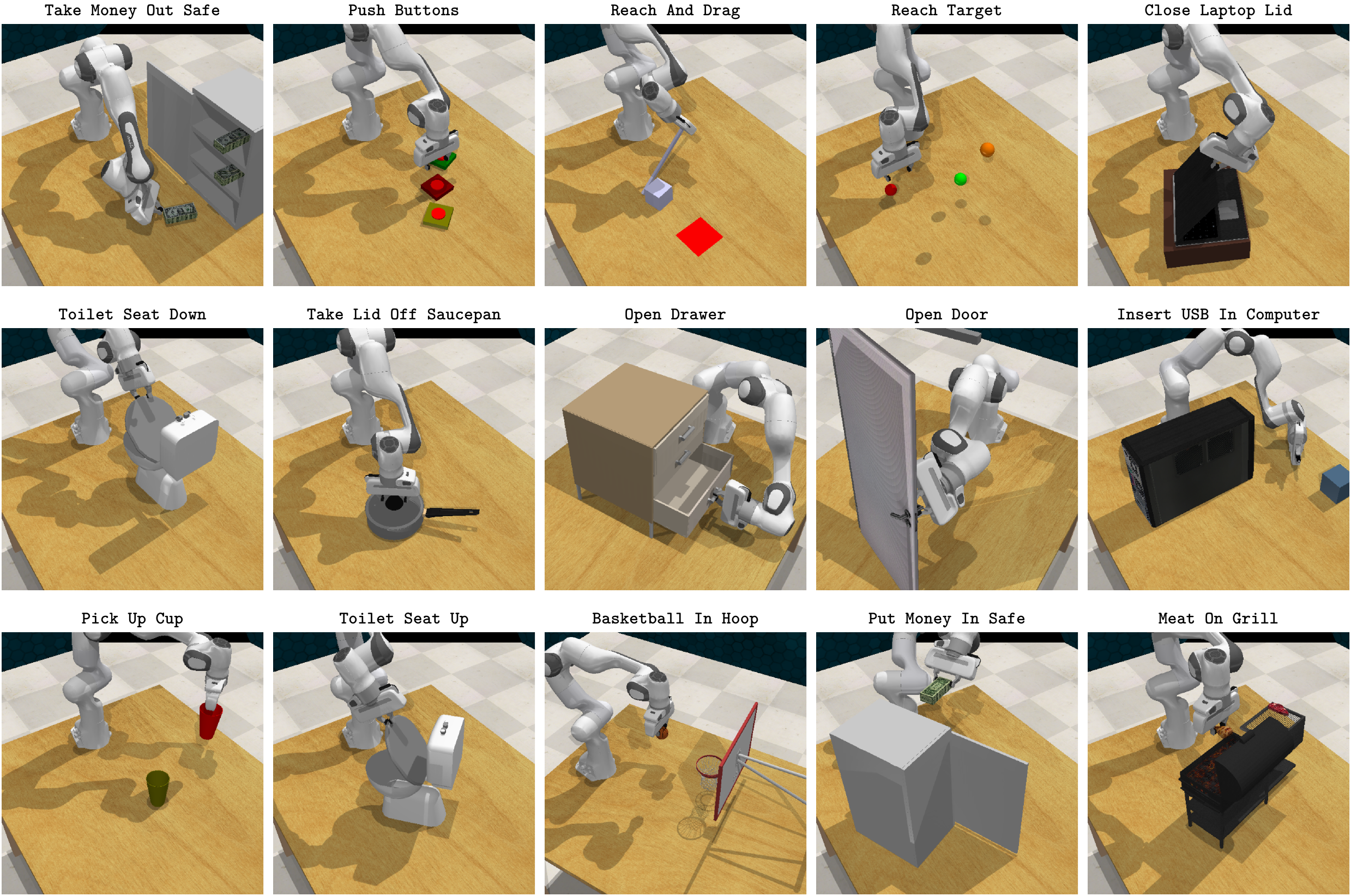}
    \captionof{figure}{Full visualization of the RLBench tasks we tested.}
    \label{fig:rlbench}
\end{figure*}

\paragraph{RLBench~\citep{james2019rlbenchrobotlearningbenchmark}.}
We use fifteen tasks from RLBench. They are simulated in CoppeliaSim through PyRep. Every task uses variation~0, so the goal is the same in every episode (e.g., always the bottom drawer, or always the red cup). The episodes still differ, because RLBench places the task at a random position on the table, with a random rotation about the vertical axis, at every reset. Several tasks also randomize individual objects; \Cref{tab:rlbench_tasks} lists these, and \Cref{fig:rlbench} shows all fifteen tasks.

\emph{Observations.} Each observation has four $84\times84$ RGB images, from the front, wrist, left-shoulder and right-shoulder cameras, stacked along the channel axis. It also has 8-D proprioception: the seven joint positions and a binary gripper-open state.
\emph{Actions.} The action is 8-D: seven delta joint positions and a discrete gripper command (open when above $0.5$). Actions are normalized to $[-1,1]$ with the per-dimension minimum and maximum of the demonstrations, widened by $20\%$ on each side.

\emph{Demonstrations.} We collect 100 demonstrations per task with RLBench's scripted motion planner. Each planner trajectory is converted to delta-joint actions, together with the next gripper state. Any demonstration with a single-step joint change above $0.2$\,rad is discarded. We do not store the planner's own observations. Instead, we replay each demonstration's actions through the training environment from the demonstration's initial scene, and store the replay, truncated at its first successful step. Replays that never succeed are discarded. Every stored transition, images included, is therefore one that the online environment itself produces.

\begin{table*}[t]
\centering
\small
\caption{The fifteen RLBench tasks and their online interaction budgets. In addition to what is listed, every task is placed at a random position and rotation in the workspace at each reset.}
\label{tab:rlbench_tasks}
\begin{tabularx}{\textwidth}{@{}>{\raggedright\arraybackslash}p{0.22\textwidth}>{\raggedright\arraybackslash}X r@{}}
\toprule
Task & Goal and scene & Online steps \\
\midrule
take money out safe   & Take the stack of money from the bottom shelf of a safe and place it on the table. The placement target is randomized. & 50k \\
push buttons          & Push the red button. Three buttons of different colors are present. & 100k \\
reach and drag        & Grasp a stick and use it to drag a cube onto the red target. & 300k \\
reach target          & Touch the red sphere with the gripper. Two distractor spheres in other colors are present, and the positions of all three are randomized. & 50k \\
close laptop lid      & Close the lid of an open laptop. & 100k \\
toilet seat down      & Lower the raised toilet seat. & 50k \\
take lid off saucepan & Lift the lid off a saucepan by its handle. & 100k \\
open drawer           & Pull open the bottom drawer of a chest of drawers. & 150k \\
open door             & Grasp the handle and push a door open. & 300k \\
insert USB in computer & Pick up a USB stick and insert it into the computer's port. & 300k \\
pick up cup           & Pick up the red cup. A second cup in a random other color is present, and the cup positions are randomized. & 200k \\
toilet seat up        & Lift the toilet seat to an upright position. & 250k \\
basketball in hoop    & Pick up a ball and drop it through a basketball hoop. & 250k \\
put money in safe     & Pick up the stack of money and place it on the bottom shelf of a safe. The position and orientation of the money are randomized. & 300k \\
meat on grill         & Pick up the chicken and place it on the grill. A steak is also present. & 200k \\
\bottomrule
\end{tabularx}
\end{table*}

\paragraph{Evaluation protocol.}
We evaluate the policy periodically on fresh episodes that are not used for training. The curves in the paper report the evaluation success rate over the course of training. The metric is the success rate: the fraction of evaluation episodes in which the success condition is reached before the episode ends. Each method is run with 3 seeds per task. For pretrained-encoder methods, seed $k$ of the online run starts from seed $k$ of the offline pretraining run. Curves show the mean success rate over seeds, smoothed with a centered moving average over 3 evaluations, or a shorter window for tasks with budgets under 300k steps. Shaded regions show the standard error over seeds.

\subsection{Model Structure Details} \label{app:model}

\paragraph{Image encoder.}
All methods except DINOv2 use the same ViT image encoder $E_\theta$, which maps the camera images to a 128-D latent $h$. The images from all cameras are stacked along the channel axis and passed through a single ViT. A small convolutional stem turns the $84\times84$ input into $9\times9=81$ tokens, which then go through the transformer blocks. The output tokens are flattened and projected to $h$ by a linear layer, a LayerNorm and a tanh. We use two sizes, listed in \Cref{tab:vit}.

\begin{table}[h]
\centering
\small
\caption{Image encoder sizes.}
\label{tab:vit}
\begin{tabular}{@{}lcc@{}}
\toprule
 & ViT-2M & ViT-20M \\
\midrule
Token width     & 128 & 384 \\
Blocks          & 1   & 8 \\
Attention heads & 4   & 6 \\
Parameters      & 1.7M & 19.7M \\
\bottomrule
\end{tabular}
\end{table}

\paragraph{Heads.}
Every head (critic, policy or value function) takes $h$ together with an embedding of the proprioceptive state from a small state encoder. Policy and value heads each have their own state encoder, while all Q-networks in the critic ensemble share a single one. All heads of an agent share the image encoder $E_\theta$. The critic is an ensemble of Q-networks, each an MLP over the image latent, the shared state embedding and the action. Following \Cref{sec:method-impl}, we treat the critic's shared state encoder as part of $f_\phi$, the first critic layer that fuses the image latent, proprioception and action.

\paragraph{Pretraining.}
Our pretraining agent consists of the image encoder, the critic ensemble, an actor and an auxiliary imitation-trained policy. The critic and actor losses train the encoder; the imitation-trained policy reads a stop-gradient copy of $h$ and does not.

\paragraph{Downstream RL.}
\begin{itemize}
    \item \textbf{EXPO}: the online agent has the image encoder, a diffusion base policy, an edit policy and the critic ensemble. Only the critic loss trains the encoder. From our pretrained model we load the image encoder $E_\theta$ and, for each Q-network, its first layer $f_\phi$ (including the critic's shared state encoder). All other parameters, including the remaining critic layers and both policies, are freshly initialized. The pretrained actor and imitation-trained policy are discarded.
    \item \textbf{PPO}: a policy head and a value head share the image encoder. From any pretrained model, PPO loads only the image encoder; its state encoders and both heads are freshly initialized.
\end{itemize}
The baselines plug a differently initialized image encoder into the same downstream agents. The IL and TCN models are pretrained on the same demonstrations, and only their image encoder is loaded. DINOv2 is a pretrained ViT-S/14 that encodes each camera separately; its outputs pass through a freshly initialized projection. The Random init baseline initializes every parameter randomly.

\subsection{Optimization Hyperparameters}
\label{sec:hp}
All learners share one optimizer setup (\Cref{tab:hp_optim}): AdamW with separate parameter groups for the encoder and the heads. The encoder uses a learning rate of $1\times10^{-4}$ with linear warmup (longer for pretrained encoders, to avoid overwriting pretrained features early) and cosine decay to zero. The heads use $3\times10^{-4}$, held constant after the same warmup. Unless a method's table states otherwise, no method tunes its own learning rate, schedule, or weight decay.
For our offline pretraining (\Cref{tab:hp_ours}), the critic adds the action-contrastive loss $\mathcal{L}_{\mathrm{AC}}$ (\Cref{eq:con_loss}) to the TD loss with weight $\alpha$ ($0.1$ on robomimic, $0.5$ on RLBench), where the logsumexp is estimated from 10 uniform actions and 10 actions each from $\pi(\cdot|s)$ and $\pi(\cdot|s')$. We run 300k updates with batch size 64 and 5 critic steps per update, using an ensemble of 10 Q-heads with the minimum over 2 heads in the TD target.
Online fine-tuning with EXPO (\Cref{tab:hp_expo}) uses the same values for every encoder initialization. The critic is updated at an update-to-data ratio of 20 on batches split evenly between demonstrations and online data. Both acting and target computation use 8 base-policy samples from a 10-step diffusion policy and 8 edits with scale 0.1, and the encoder is trained by the critic loss only.
PPO (\Cref{tab:hp_ppo}) first warm-starts the policy with 50k steps of imitation learning. This warm start also trains the encoder for every initialization (including Random init), though for far fewer steps than encoder pretraining (300k offline updates for \methodname). PPO then runs 300, 200, and 150 updates on \emph{square}, \emph{can}, and \emph{lift}, respectively. Each update collects 64 episodes and trains for 10 epochs with a fixed policy standard deviation of 0.1 and no entropy bonus. PPO uses its own actor and critic learning rates, and unlike in EXPO, its encoder is trained by both the policy and value losses.

\begin{table}[h]
\centering
\small
\caption{Optimizer shared by all learners.}
\label{tab:hp_optim}
\begin{tabularx}{\linewidth}{@{}>{\raggedright\arraybackslash}p{0.36\linewidth}>{\raggedright\arraybackslash}X@{}}
\toprule
Optimizer               & AdamW \\
Encoder learning rate   & $1\times10^{-4}$, linear warmup, cosine decay to 0 \\
Encoder warmup          & 15k steps (pretrained), 5k steps (from scratch) \\
Encoder weight decay    & 0.01 \\
Head learning rate      & $3\times10^{-4}$, same warmup, then constant \\
Head weight decay       & $1\times10^{-4}$ \\
Temperature             & Adam, lr $3\times10^{-4}$, initial value 0.1 \\
\bottomrule
\end{tabularx}
\end{table}

\begin{table}[h]
\centering
\small
\caption{Hyperparameters of our offline pretraining.}
\label{tab:hp_ours}
\begin{tabularx}{\linewidth}{@{}>{\raggedright\arraybackslash}p{0.42\linewidth}>{\raggedright\arraybackslash}X@{}}
\toprule
Offline updates                         & 300k \\
Critic steps per update                 & 5 \\
Batch size                              & 64 \\
Q ensemble size / heads in target min   & 10 / 2 \\
Discount $\gamma$                       & 0.99 \\
Target update rate $\tau$               & 0.005 \\
Actor target entropy                    & $-|\mathcal{A}|/2$ \\
$\mathcal{L}_{\mathrm{AC}}$ weight $\alpha$            & 0.1 (robomimic), 0.5 (RLBench) \\
Logsumexp samples per state             & 10 uniform, 10 from $\pi(\cdot|s)$, 10 from $\pi(\cdot|s')$ \\
Logsumexp temperature                   & 1 \\
\bottomrule
\end{tabularx}
\end{table}

\begin{table}[h]
\centering
\small
\caption{Hyperparameters of online EXPO (all initializations).}
\label{tab:hp_expo}
\begin{tabularx}{\linewidth}{@{}>{\raggedright\arraybackslash}p{0.42\linewidth}>{\raggedright\arraybackslash}X@{}}
\toprule
Update-to-data ratio (critic)           & 20 \\
Base-policy steps per update            & 2 \\
Batch size                              & 64 (50\% demos, 50\% online) \\
Diffusion steps                         & 10 \\
Base samples / edits (acting and target) & 8 / 8 \\
Edit scale                              & 0.1 \\
Q ensemble size / heads in target min   & 10 / 2 \\
Discount $\gamma$                       & 0.99 \\
Target update rate $\tau$               & 0.005 \\
Edit-policy target entropy              & $-|\mathcal{A}|/2$ \\
Encoder trained by                      & critic loss only \\
\bottomrule
\end{tabularx}
\end{table}

\begin{table}[h]
\centering
\small
\caption{Hyperparameters of PPO (all initializations).}
\label{tab:hp_ppo}
\begin{tabularx}{\linewidth}{@{}>{\raggedright\arraybackslash}p{0.42\linewidth}>{\raggedright\arraybackslash}X@{}}
\toprule
IL warm start                           & 50k steps, batch size 256 (trains policy and encoder) \\
PPO updates                             & 300 (Square), 200 (Can), 150 (Lift) \\
Episodes per update                     & 64 \\
Epochs per update                       & 10 \\
Minibatch size                          & 12{,}800 transitions \\
Clip ratio                              & 0.1 \\
Discount $\gamma$ / GAE $\lambda$       & 0.99 / 0.95 \\
Value loss coefficient                  & 0.5 \\
Entropy coefficient                     & 0 \\
Max gradient norm                       & 1.0 \\
Policy standard deviation               & 0.1 (fixed) \\
Actor / critic learning rate            & $3\times10^{-4}$ / $1\times10^{-3}$ \\
Encoder trained by                      & policy and value losses \\
\bottomrule
\end{tabularx}
\end{table}

\subsection{Baseline Implementation Details}
All baselines use the same downstream agents as \methodname. The implementation details for the different objectives are as follows:

\paragraph{Random init.}
The ViT is randomly initialized and receives no representation pretraining. It is trained from scratch, together with the rest of the agent, by the downstream algorithm; for PPO, this includes the same 50k-step IL warm start that every initialization receives (\Cref{sec:hp}). This baseline measures what online RL achieves without any representation pretraining.

\paragraph{Imitation-pretrained encoder.}
The encoder is pretrained by imitation learning with a diffusion policy. The diffusion policy, with the same architecture as EXPO's base policy, is conditioned on the image latent $h = E_\theta(o)$ and the proprioceptive state $q$. It is trained with the standard DDPM objective of predicting the noise added to a demonstration action:
{\small
\begin{equation*}
\begin{split}
\mathcal{L}_{\mathrm{IL}} = \mathbb{E}_{(o,q,a)\sim\mathcal{D},\,k,\,\epsilon\sim\mathcal{N}(0,I)}
\Big[ \big\| \epsilon - \epsilon_\omega\big(&\sqrt{\bar\alpha_k}\,a + \sqrt{1-\bar\alpha_k}\,\epsilon,\\
&k,\; E_\theta(o),\; q\big) \big\|^2 \Big]
\end{split}
\end{equation*}
}
where $k$ is the diffusion step. The gradient of this loss trains both the denoiser $\epsilon_\omega$ and the encoder $E_\theta$. The encoder therefore learns features that predict the demonstrated actions, but it receives no reward or value signal.

\paragraph{Time-contrastive encoder~\citep{TCN2017}.}
Following R3M~\citep{pmlr-v205-nair23a}, we pretrain the encoder with a time-contrastive objective. R3M's other terms require language annotations, which our demonstrations do not have, so we use only the time-contrastive term. The objective uses neither actions nor rewards. From one demonstration, we sample three frames at times $t_0 < t_1 < t_2$, so that frame $t_1$ is closer in time to each endpoint than the endpoints are to each other. Similarity is the negative squared distance between latents, $S(x, y) = -\|E_\theta(x) - E_\theta(y)\|_2^2$. Each endpoint serves as an anchor in an InfoNCE loss. The positive is frame $t_1$. The negatives are the other endpoint and frames $y_n$, $n = 1, \dots, N$, from other demonstrations (we use $N=3$):
\begin{equation*}
\mathcal{L}_{2} = -\log \frac{e^{S(o_{t_2}, o_{t_1})}}{e^{S(o_{t_2}, o_{t_1})} + e^{S(o_{t_2}, o_{t_0})} + \sum_{n} e^{S(o_{t_2}, y_n)}},
\end{equation*}
and $\mathcal{L}_0$ is defined symmetrically with anchor $o_{t_0}$. The total loss is $\mathcal{L}_{\mathrm{TCN}} = (\mathcal{L}_0 + \mathcal{L}_2)/2$. It pulls temporally close frames of the same trajectory together and pushes apart temporally distant frames and frames from other trajectories. The loss is applied directly to the latent $h$ that the downstream agent loads, so no separate projection head is trained and then discarded.

\paragraph{DINOv2 encoder~\citep{oquab2024dinov2learningrobustvisual}.}
We use the publicly released DINOv2 ViT-S/14, pretrained with self-supervised learning on a large curated set of natural images. It is not trained specifically on robot data. To show it the same pixels as every other encoder, we apply it to the $84\times84$ frames without resizing. This gives a $6\times6$ grid of patches per camera, and we interpolate the pretrained positional embeddings to that grid. Following the standard ImageNet preprocessing, the frames are normalized with the ImageNet mean and standard deviation. One set of DINOv2 weights encodes each camera separately. For each camera we take DINOv2's standard linear-probe feature: the CLS token concatenated with the mean of the patch tokens. These features are concatenated across cameras and mapped to the 128-D latent $h$ by a freshly initialized projection (a linear layer, a LayerNorm and a tanh). During online RL, the DINOv2 weights are fine-tuned along with the rest of the agent.

\end{document}